\documentclass[a4paper,10pt]{article}
\usepackage[super,square]{natbib}
\usepackage{amsfonts}
\usepackage{float}
\usepackage{graphicx}
\usepackage{subcaption}
\usepackage{amsmath}
\DeclareMathOperator*{\argmax}{arg\,max}

\usepackage{bbm}
\usepackage{graphicx}
\graphicspath{ {./images/} }
\usepackage{anysize}
\marginsize{2cm}{2cm}{1cm}{2cm}
\usepackage{fancyhdr}

\usepackage{soul}
\usepackage[dvipsnames]{xcolor}
\usepackage{hyperref}
\hypersetup{
    colorlinks=true,
    linkcolor=black,
    citecolor=CadetBlue,
    filecolor=CadetBlue,      
    urlcolor=CadetBlue,
}
\usepackage{lipsum}
\usepackage{multicol}
\renewenvironment{abstract}
 {\par\noindent\textbf{\abstractname}\ \ignorespaces \\}
 {\par\noindent\medskip}

\begin{document}
\pagestyle{fancy}
\thispagestyle{empty}
\fancyhead[L]{}
\renewcommand*{\thefootnote}{\fnsymbol{footnote}}
\begin{center}
\Large{\textbf{Computing the optimal keyboard through a geometric analysis of the English language}}
\vspace{0.4cm}
\normalsize
\\ Jules Deschamps, Quentin Hubert, Lucas Ryckelynck  \\
\vspace{0.1cm}
\textit{Columbia University,}
\small{New York}
\medskip
\normalsize
\end{center}
{\color{gray}\hrule}
\vspace{0.4cm}
\begin{abstract}
In the context of a group project for the course \textbf{COMSW4995 002 - Geometric Data Analysis}, we have brought our attention to the design of fast-typing keyboards. Leveraging some geometric tools in an optimization framework allowed us to propose novel keyboard layouts that offer a faster typing.
\end{abstract}
{\color{gray}\hrule}
\medskip

\tableofcontents
\section{Introduction}
\subsection{Motivation}
Most keyboard layouts such as QWERTY are designed so that the whole keyboard is used homogeneously \cite{Noyes1982}, which was particularly useful when typewriters were the only way to print text efficiently. The whole idea of QWERTY is about spreading the most often used characters over the keyboard space in order to avoid malfunctions. This means this layout is not optimized in terms of typing time and yet it is still widely used nowadays by convention. However, the issues it addressed back then are no longer and the non-optimality of this layout is an issue for digital devices. For example, writing on a phone with one finger is harder than it should be with QWERTY.

In this regard, here we propose novel keyboard layouts that address this particular issue of reducing the writing time.

\subsection{Relative Work}
Designing an optimal keyboard is a recurrent interest for optimization researchers. However, the definition of \textit{optimal} often varies, as well as the evaluation metrics and of course the models. Indeed, modeling how one types on a keyboard demands many assumptions about physical behaviors, especially when two hands are considered. Furthermore, evaluation metrics mostly rely on a text corpus, but the hyper-parameters of the model, as well as the representativity of the evaluation text have to be studied. We also have to be aware that it is a very applied problem, meaning that each article tackling the problem does not inspire his proposition to the problem from other relative to the same topic, but are mostly applications for cutting edge methods developed for other purposes. Hence, there is no continuity in the litterature of this famous problem.

In \cite{genetic2020}, Amir H. H. Onsorodi and Orhan Korhan model the optimal keyboard problem into an Integar Linear Programming, and use a genetic algorithm to approximate a layout solution, instead of needing huge computation with exact Branch$\&$Bound.

In \cite{ant2003}, Reearchers from \textit{Centrale Paris} model the optimal keyboard as a two-cluster object, and define an objective function which is the linear combination of different sensitivities to some metrics (not using too much a finger, etc.). Finally, they solve the Integer-Programming through a Ant-colony algorithm, which is a kind of probabilistic Genetic Algorithm.

In \cite{swarm2011}, they use a very promising method, which learns from data. This consists in simulating typing, with a stochastic algorithm predicting  the next word coming after, and updating layout of the keyboard at each iteration.

Most of these research works do not elaborate about typing assumptions and the geometry of their final solution. Hence our main contributions are
\begin{itemize}
    \item Modeling English Language wording through Markov Chain and using its long-run distribution
    \item Using a ILP-cluster approach, along data-based Markov transitions, to solve the layout
    \item Analyzing the geometry of our solution, with curvature and covariance ellipses
\end{itemize}

\subsection{Modeling}
The time spent writing is directly related to the distance covered by the hands, as we consider the speed of hand to be constant. It is worth mentioning that this relationship is different for 1-handed and 2-handed typing. We will be mindful of that, while maintaining our focus on reducing the distance travelled by the hand(s).

Our problem is then to find out the optimal collection of coordinates $X$ of the 26 letters composing the Latin alphabet, through minimizing the travelled distance by our hands. We will solve this problem thanks to a training dataset of the most used words in English, allowing to compute the frequency of the transition from letter $i$ to letter $j$.

\newpage

{\color{gray}\hrule}
\begin{center}
\section{Pre-processing}
\end{center}
{\color{gray}\hrule}
\vspace{10pt}

\subsection{Distribution of the English language}
The English language is composed of words, encoded in letters. In the context of designing the optimal keyboard, the frequency of use of each letter, along with the transition probabilities from one letter to another, seem to be relevant.

We obtain this data in two steps:
\begin{itemize} 
\item from a dataset of English words and their frequency accessible \href{https://www.kaggle.com/datasets/rtatman/english-word-frequency}{here}, we extract the instances of 2-letter sequences (for example, in \textit{'the'} we find the sequences \textit{'th'} and \textit{'he'}) and the number of occurrences for every $i \rightarrow j$;
\item we input these occurrences in a $26 \times 26$ count matrix, where the $(i,j)$ entry corresponds to $i \rightarrow j$.
\end{itemize}
\subsection{Markov Chain Modeling}
We now set a mathematical framework for our study. We define the state space $\{a,b,...z\}$ as the letter we are currently typing. We make the assumption that being on a state $i$ is only dependent on the previous state, and we note $P_{ij}$ be the probability of transitioning from letter $i$ to letter $j$, by normalizing each row of the count matrix.  

Hence, we can define a Markov Chain, with probability matrix $P$ and stationary distribution $\pi$ such that $\pi = \pi P$ (this distribution is in fact the overall frequency of each letter and can be retrieved directly from the original dataset).

\begin{figure}[h!]
    \centering
    \begin{subfigure}[b]{0.45\textwidth}
        \includegraphics[width=\textwidth]{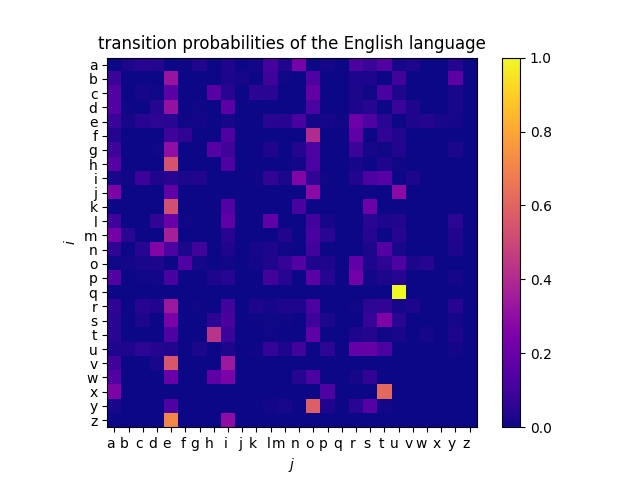}
        \caption{Transition matrix of $X$: $P_{ij} = \mathbb{P}(i \rightarrow j)$}
    \end{subfigure}
    \hfill
    \begin{subfigure}[b]{0.45\textwidth}
        \includegraphics[width=\textwidth]{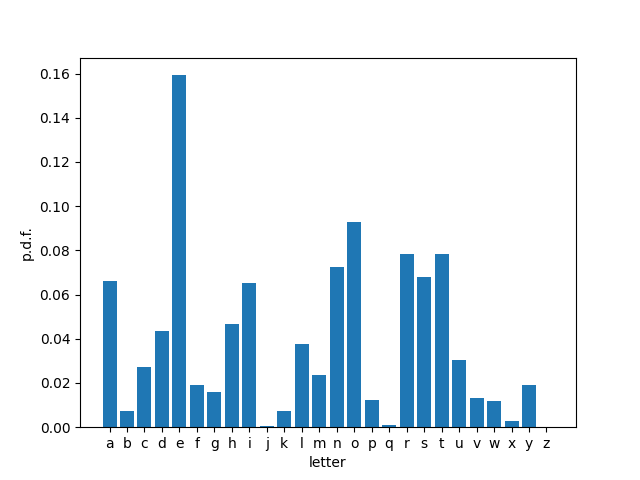}
        \caption{$\pi$: the distribution of the alphabet in the English language}
    \end{subfigure}
\end{figure}

This set-up of the problem is a graph approach, $G = (V,E)$, being respectively the vertex and the edge sets, $P$ being the weighted adjacency matrix of the graph (i.e., the transition matrix of the associated Markov Chain) and the problem consists in finding out an optimal way to compute its geometry to reduce the traveled distance as much as possible.

\newpage
{\color{gray}\hrule}
\begin{center}
\section{Geometric non-layout approach}
\end{center}
{\color{gray}\hrule}

\vspace{10pt}

This section aims at uncovering the geometry of the optimal keyboards, without some of the layout constraints (i.e., the physical constraints). In that regard, we see letters as points, and the problem becomes that of finding the best setting of these letters in $\mathbb{R}^2$ so as to reduce the time spent typing. \\
From now on, we will always keep 2 approaches, for 2 different keyboards: the 1-handed keyboard (H1), and the 2-handed keyboard (H2). Both can be useful in different cases: indeed, although using 2 hands is largely more popular, typing on one's smartphone is also accessible with 1 hand only (with SwiftKey for example, you can simply slide your finger through the letters). Also, we don't consider the \textit{space} bar or any punctuation key but only letters. This is an evident oversimplification of the problem at hand but it uncovers some interesting properties about the English language.
\medskip

\subsection{First approach: one-handed typing}
\subsubsection{Optimization problem}
As stated in 2), we note $X$ the Markov Chain associated with the letter one is typing, we note $P_{ij}$ the probability of going from $i$ to $j$ (i.e., typing $j$ after $i$) in the English language, $d_{ij}$ the distance between keys $i$ and $j$, and $\pi_i$ the long-run proportion spent on $i$.
For all time $t$ we look to minimize the average distance covered by the hand, which is the quantity (we sum only for $i \neq j$ because all $d_{ii}$ must be 0): $$\frac{1}{t} \sum_{1 \leq i \neq j \leq n} \sum_{k=0}^{t-1} \mathbbm{1}(X_k = i \cap X_{k+1}=j)d_{ij} = \frac{1}{t} \sum_{1 \leq i \neq j \leq n} \sum_{k=0}^{t-1} \mathbbm{1}(X_k = i) \mathbbm{1}(X_{k+1} = j | X_k = i) d_{ij}.$$

The Law of Large Numbers and the Markov property ensure that this quantity converges to its mean, i.e.: $$\sum_{1 \leq i \neq j \leq n} \pi_i P_{ij} d_{ij}.$$

Therefore, under a coherent set of constraints over the $d_{ij}$, this is the quantity that we will minimize, which we will note $L(d)$ from now on.

We propose the following constraints, with hyperparameters $d_{min}$ and $b$:
$$\begin{cases} \forall i, j, d_{ij} =  d_{ji} \text{ (the distance matrix is symmetric)} \\ \forall i, j, d_{ij} \geq d_{min} > 0 \text{ (keys must be distant by a minimal distance, their size at least)}\\ 
||d||_2 \geq c > 0 \text{ (scale constraint)} \end{cases}$$

However, this problem can be greedily solved without regularization: indeed it suffices to set all distances to $d_{min}$ in a first place, and then to consider the pair(s) $i, j$ for which $\pi_i P_{ij} + \pi_j P_{ji}$ is minimized, and set the corresponding distance $d_{ij}$ to the minimal quantity to satisfy the second constraint (this quantity $d$ satisfies $d^2 + (\frac{n(n-1)}{2} - 1) d_{min}^2 = c^2$). As we want to limit the sparsity of our distance matrix, Ridge regularization seems fit to integrate in our problem.

In the end, the problem we want to solve is:

$$\min_{d = (d_{ij}, 1 \leq i \neq j \leq n)} L(d) + \alpha ||d||_2^2$$
$$\text{s.t. } \begin{cases}  \forall i, j, d_{ij} =  d_{ji} \\ \forall i, j, d_{ij} \geq d_{min} > 0 \\ 
||d||_2 \geq c > 0 \end{cases}$$

We will solve for an array of hyperparameters to find the ones we deem the most fit for our application. We shall mention that $c$ and $d_{min}$ both refer in a sense to the scale of the keyboard so we can set set $c=1$ and then reduce to 2 hyperparameters ($d_{min}$ and $\alpha$).

This allows to metrize the alphabet in the English language.

\subsubsection{From distances to points}

The optimization process provides us with a distance matrix between all letters. We now want to uncover the geometry of the metric space that this induces. For that, we embed our $N=26$ points into $\mathbb{R}^2$ using Multi Dimensional Scaling (MDS):
\begin{itemize} 
\item we transform the distance matrix $D = (d_{i,j})_{i,j}$, into a Gram matrix $G = (<x_i,x_j>)_{i,j}$, using double centering;
\item we keep the first 2 eigenvectors associated with the highest eigenvalues, and we project our space on the space spanned by these vectors. 
\end{itemize}

The outcome is a matrix of $N$ points in $\mathbb{R}^2$, which is the distribution of the keys of the keyboard. This is the result of an isometry, so the distribution is not directly implementable on a keyboard, as it does not form a rectangle with equal distance between neighbors. But the relaxed solution can then be modified to meet these requirements (See Part 4).

\subsubsection{Results and limits}
\includegraphics[scale=.5]{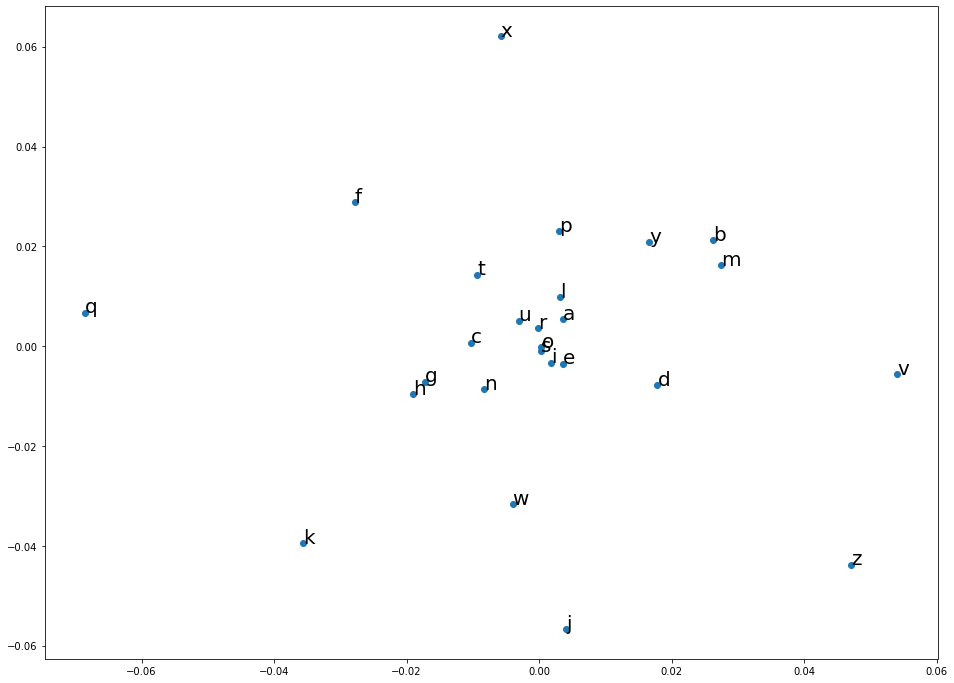}

This is with $d_{min}=0.01$, $c=1$, $\alpha=1.$ Hyperparameters are subject to change but we have found that this gave a sparse enough embedding of the points in $\mathbb{R}^2$. For that reason we will use the same one when using a similar optimization framework, on a subset of the alphabet.

The results are expected: letters most frequently used are at the center of the embedding and the less frequent a letter seems to be used the further it gets from the center. For instance, letters like \textit{q, z} delimit the frontier of the embedding. It is hard at this stage to find an interpretation at the transition level however (i.e. how much the $\pi_i$ and the $P_{ij}$ altogether influence the MDS embedding).

All this is merely to implement a framework to represent the language as a keyboard. However, this keyboard serves no interpretation: in a sense, it is only a lower-resolution representation of the distances that we have computed through optimization.

\subsection{More accurate approach: two-handed typing}
In the previous subsection, we modeled the problem as a minimization of the overall travelled distance. However, a more realistic approach is to consider that we type on a keyboard with two hands.
\subsubsection{Well-posed problem}

Writing with 2 hands on the keyboard largely refines the problem at hand.
In the following we will consider that the keyboard consists of 2 clusters only, one for each hand, and that the left keyboard is inaccessible to the right hand and vice versa.
As before, we wish to minimize the time spent writing on the keyboard, and therefore the sum of the distances covered by each hand. Now, as opposed to our first approach in which we only considered $1$-transitions, we have to consider all $k$-transitions here: to be clear, if we want to write the word \textit{andrew} and that the letters \textit{a} and \textit{w} only are on the right keyboard, while the others are on the left keyboard, we have to take the distance $d_{aw}$ into account (as it is the distance covered by the right hand here), pondered by the probability of having such a 5-transition.

In math terms, this means that for a given clustering (partition) of our 26 letters into clusters $A$ and $B$, we have to ponder the distance $d_{ij}$ where $i,j \in B$ by the probability of typin only letters on cluster A during the $i \rightarrow j$ transition, i.e.: $$\sum_{k} P_{iA} P_A^k P_{Aj} = P_{iA} (I - P_A)^{-1} P_{Aj},$$where: $$(P_A)_{kl} = \begin{cases} P_{kl} \text{ if } k, l \in A \\ 0 \text{ otherwise} \end{cases}.$$

In the end, the new objective is: $$ \min_{A,B,d} \sum_{i,j \in A} \pi_i(P_{ij} + P_{iB}(I-P_B)^{-1}P_{Bj})d_{ij} + \sum_{i,j \in B} \pi_i (P_{ij} + P_{iA}(I-P_A)^{-1}P_{Aj})d_{ij}.$$

But, that is practically intractable given the complexity of computing the $(I-P_A)^{-1}$. We have tried to come up with approximations but they are either too expensive too or they change the nature of the problem too much.

\subsubsection{Relaxation: clustering and two-fold optimization}

We can relax the problem by decomposing it as a two-step problem : first finding out an optimal cut on the keyword ($k$ letters on a side and $n-k$ on the other) and then minimizing distances covered in both parts.

Our intuition here is to maximize inter-cluster transitions as they might be considered as relatively instant. Whereas an intra-cluster transition will need for the hand to move from on key to the other within the cluster, the time needed for an inter-cluster transition will not count the repositioning time spent by the hand on the other cluster: when writing the word \textit{cat}, if \textit{c, a} are in the right cluster and \textit{t} in the left cluster, chances are the transition from \textit{a} to \textit{t} will be swift because the left hand will have had time to position itself before the right hand types \textit{a}.

With the same notations as before, introducing $f$ as a regularization function and $A,B$ a partition of the alphabet, we solve :
$$\{A, B\} = \argmax_{A,B} (\pi_A P_{AB}\mathbf 1 + \pi_B P_{BA}\mathbf 1)\times \left(\frac{1}{f(\pi_A \mathbf 1)}+\frac{1}{f(\pi_B \mathbf 1)} \right) $$
$\mathbf{1}$ is a notation for summing all coefficients of the vector. Empirically, we have found that $x \geq 0 \mapsto f(x) = \sqrt{x}$ is both simple enough and gives balanced enough clusters to choose it here. This is a non-linear integer programming problem, which can be solved by brute-force ($2^{26} \approx 7 \times 10^7$ possibilities).

The second step is then to apply separately our solution of Problem 3.1.2 to the 2 clusters of letters we obtained, to have the desired distances and then the associated points.
\newline

The implementation reads:

$\textbf{Step 1} : x_A = \text{diag}((\delta_{iA})_i), x_B = \text{diag}((\delta_{iB})_i) $ 
$$\{A, B\} = \argmax_{A,B} (\pi x_A P x_B\mathbf 1 + \pi x_B Px_A\mathbf 1)\times \left(\frac{1}{\sqrt{\pi x_A \mathbf 1}}+\frac{1}{\sqrt{\pi x_B \mathbf 1}} \right)$$

$\textbf{Step 2} : $
$$\min_{d_i} L(d_i) + \alpha||d_i||^2_2 \text{ for $i \in A$ and then $i \in B$, separately.} $$

We get the following results.

\begin{figure}[H]
    \centering
   \includegraphics[scale=.4]{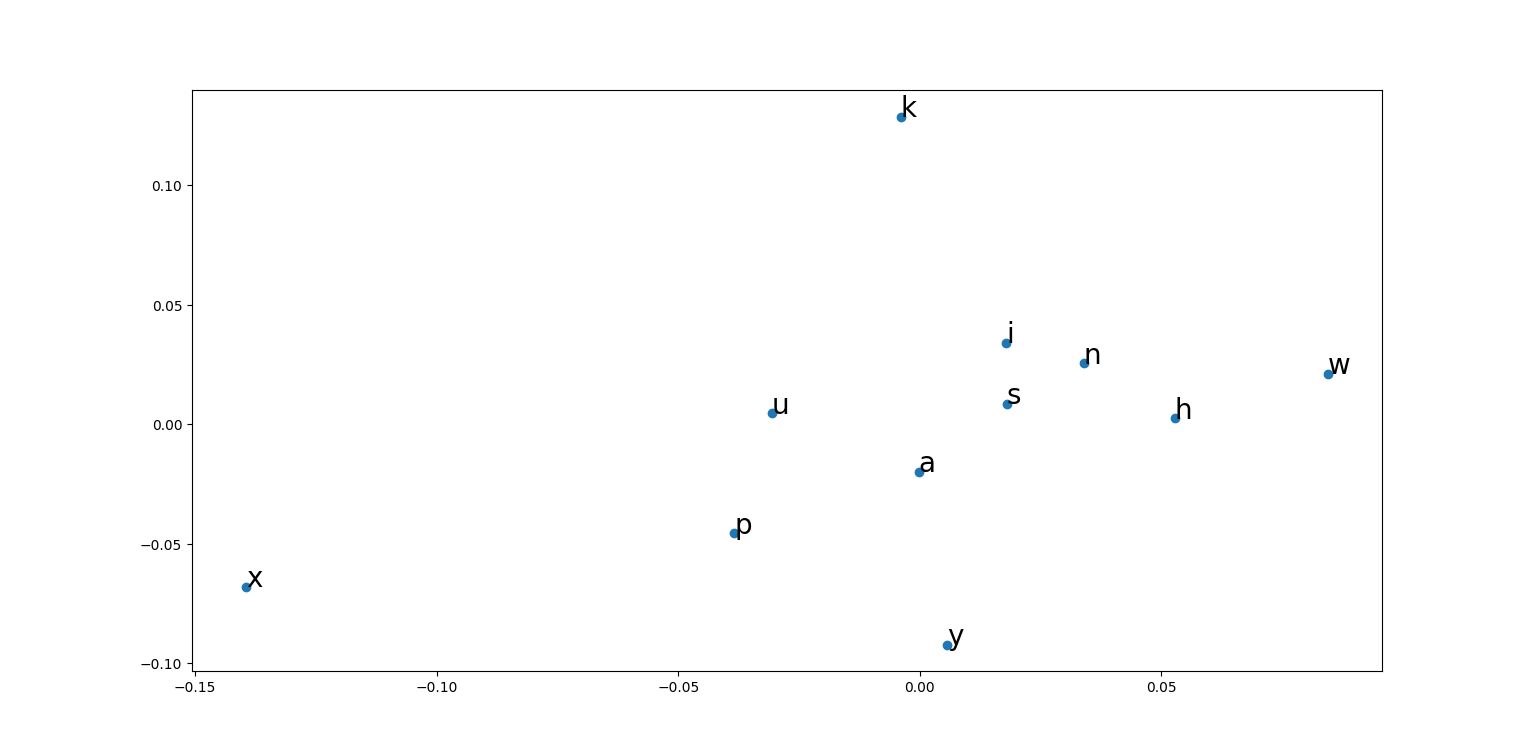}
   \caption{2D-embedding of the first cluster}
    \includegraphics[scale=.4]{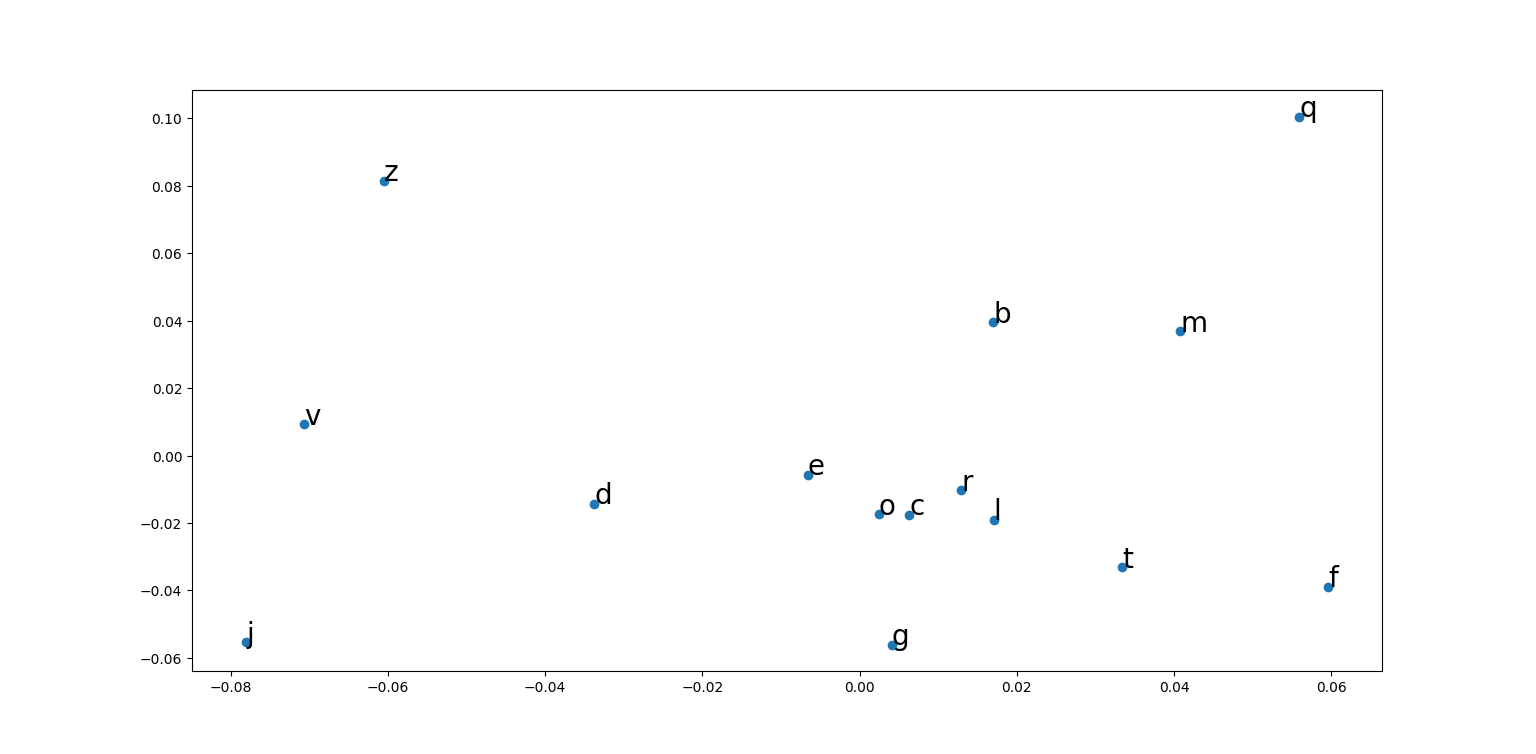}
    \caption{2D-embedding of the second cluster}
\end{figure}

We draw a similar interpretation that for the 1-handed keyboard: except for saying that a letter has more 'gravity' as it is frequently used (and thus is closer to the center of the embedding), it is hard to understand the influence at the transition level.

\subsection{Curvature analysis}

In this section we conduct an analysis on the curvature of the keyboard arrangements we have obtained so far, i.e. on the 1-handed keyboard, the 2-handed one..

Measuring the curvature from the keyboards we have obtained should extract a lot of information on how they should be used, and, at large, should give a perspective on how the English language is used in terms of letters. The main notion of curvature we can use to explore that information is discrete Ricci curvature, which is fitting in the case of a graph. The more an edge is "congested" with transitions from one end (letter) to the other, the more negative the curvature is along that edge. As such, an edge with positive curvature can be interpreted as more of an "outskirt" edge of the graph.

The discrete Ricci curvature measures the curvature of an edge in a graph. We model our keyboard as a weighted graph $G = (V, E)$, where $V$ is the alphabet and $E$ is the set of all edges, weighted by their associated distances obtained throughout our previous analyses. Let $k > 0$, $(x, y) \in E$, we take discrete Ricci curvature along the direction $xy$ to be: $$\kappa_{xy} = 1 - \frac{d_W(\mu_x, \mu_y)}{d_{xy}},$$where $\mu_x$ is the distribution on $kNN(X)$ (the $k$ nearest neighbours of $x$, including itself), extracted from the Markov Chain, and $d_W$ is the Weierstrass distance.\\
To be clear, for $i\in V $ we have: $$\mu_x(i) = \begin{cases} \frac{P_{xi}}{\sum_{j \in kNN(x)} P_{xj}} \text{ if } i \in kNN(X) \\ 0 \text{ otherwise} \end{cases}.$$

We extend discrete Ricci curvature to discrete Gauss curvature to extract local curvature for each point: $K_x = \kappa_{x,max} \times \kappa_{x,min}$. This will allow for a more individual interpretation of the keyboards as sets of letters.

We shall note that all these quantities depend on the value of $k$ chosen at the beginning of this subsection. To offset the dependency on $k$ that might be too high sometimes, we will average curvatures over an array of values of $k$. However, as the graphs we consider are quite connected, it might seem misguided to account for too many nearest neighbors.

\subsubsection{H1 keyboard}

Here we analyze the curvature of the alphabet in the English language. We are able to do so because we derived a metric on the alphabet in earlier sections.

\begin{figure}
    \centering
    \includegraphics[scale=.4]{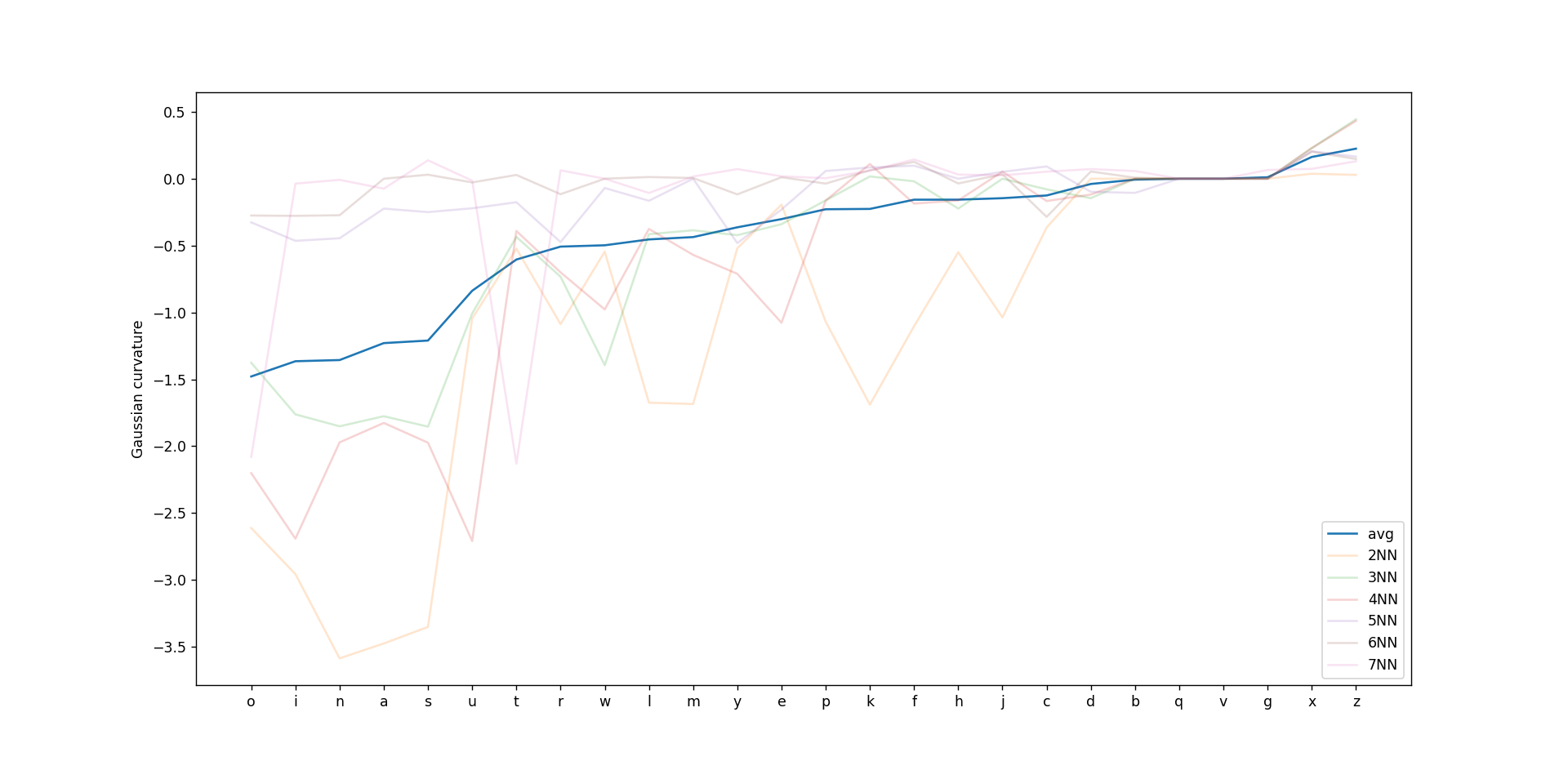}
    \caption{Discrete Gaussian curvature of each letter of the alphabet in the English language. \\ $k$NN is computed for $k \in 2:7$ and the \textbf{blue} curve is the average Gaussian curvature.}
\end{figure}

We notice that the ranking of the alphabet in terms of curvature has some similarities with the frequency of the letters in the English language: letters like \textit{z}, which are sparsely used, tend to have greater curvature, than letters that are most used, such as \textit{o} or \textit{e}. This follows the interpretation that transitions from or to \textit{o} are more 'congested' than the ones with \textit{z}. There are some dissimilarities though: for instance, \textit{c} is the 10th most used letter, while it is the 8th most curved letter - conversely, \textit{w} is the 7th least used letter but it ends up with the 9th lowest curvature. As such, we could consider that the letters with highest curvature are the least useful in the English language.

We note that almost all letters are negatively curved: this is closely related to the fact that a lot of them find a path from one to another. Some stand out, though: in particular, \textit{z} and \textit{x} have 'somewhat-positive' curvature. This is not that surprising, considering almost all transitions taking these letters into account are univocal (see the transition probabilities), as in almost all occurences of \textit{x} are at the end of a word. Intuitively, from that observation alone, \textit{x} should be located at the 'extremity' of the alphabet, hence the positive curvature. In that sense, letters with positive curvature are the least useful in the English language and should be the ones to be discarded if the alphabet were to shrink.

\subsubsection{H2 keyboard}

\begin{figure}[H]
    \centering
    \includegraphics[scale=.4]{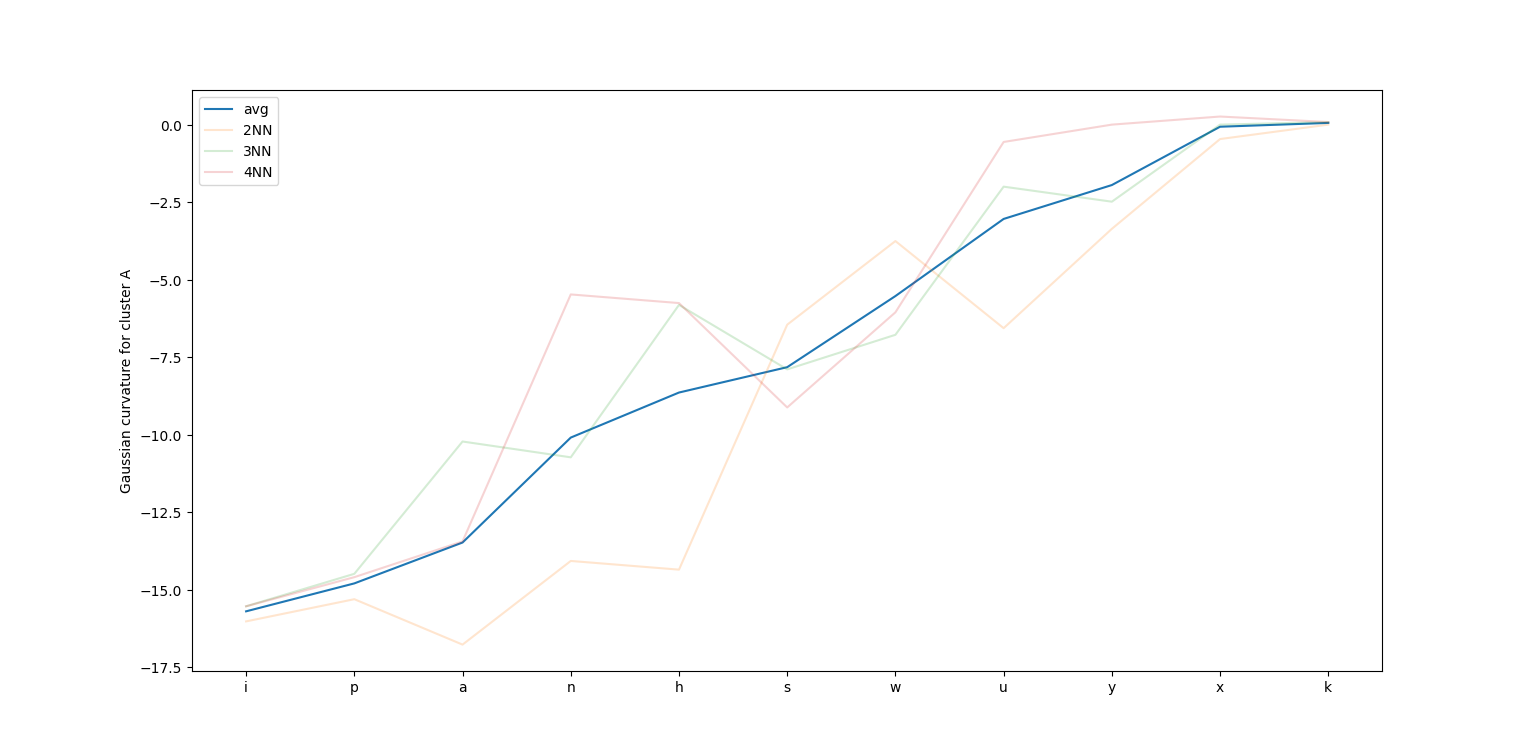}
    \caption{Discrete Gaussian curvature of each letter in the first cluster}
    \includegraphics[scale=.4]{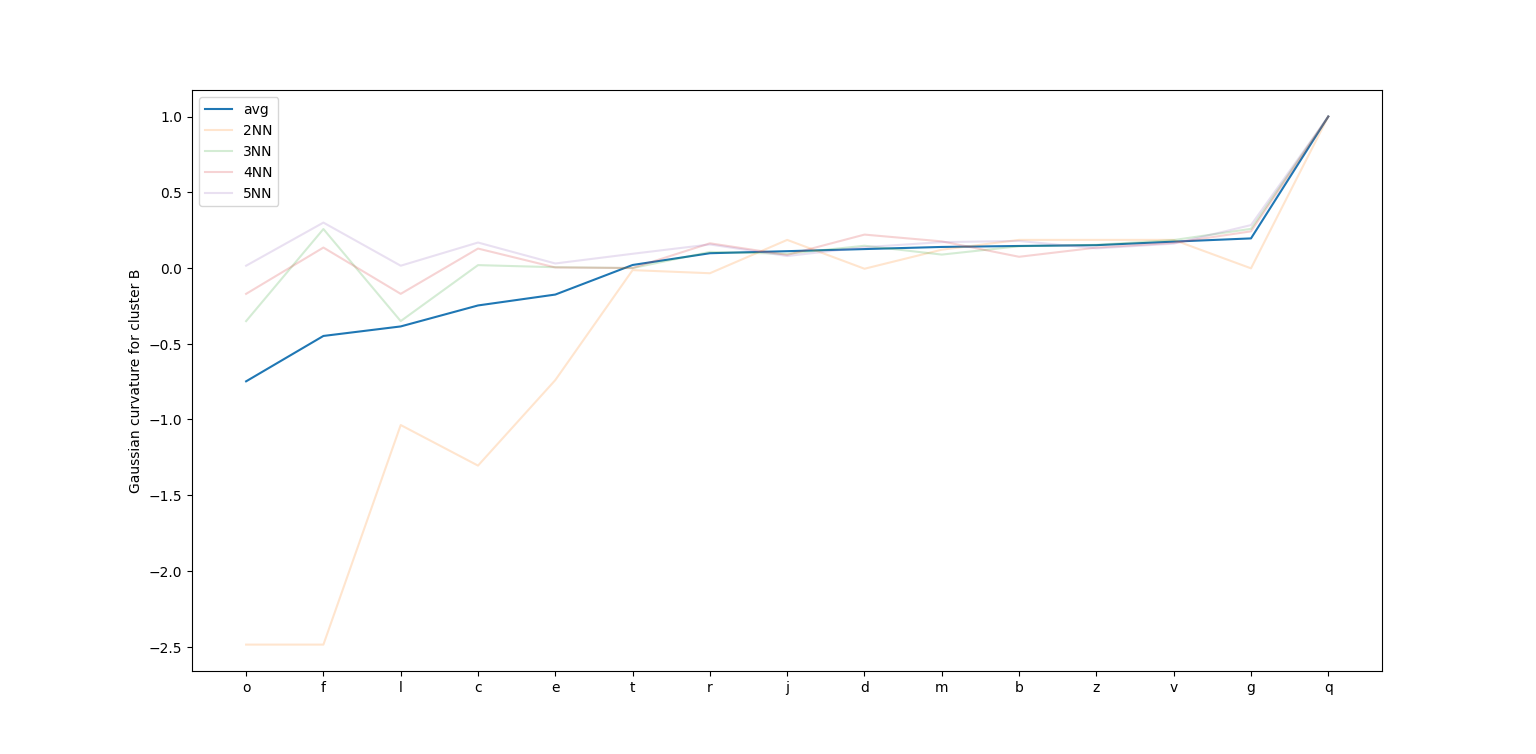}
    \caption{Discrete Gaussian curvature of each letter in the second cluster}
\end{figure}

Here, we conduct the curvature analysis on each sub-keyboard of the H2 keyboard (i.e., the left and right parts).

Similarly, we conclude that the letters \textit{k, x, q, g} and others should end up on the sides of each sub-keyboard.

It is interesting to note that the first clustering is a lot more negatively curved than the second one: while the first one is distributed quite uniformly between $-16$ and $0$, the second cluster plateaus around $0$ and ranges from $-0.7$ to $1$. The second cluster is relatively flat while the first one could be embedded in the hyperbolic space.

This might stem from the clustering: the quantity we have looked to maximize may be closely related to curvature in a sense. However we haven't been able to explicit a link so far.

\vspace{10pt}
\newpage

{\color{gray}\hrule}
\begin{center}
\section{Layout approach: integer programming}
\end{center}
{\color{gray}\hrule}

\vspace{10pt}

In this section we look to embed our alphabet in a keyboard layout directly with integer programming.

A layout can simply be seen as a subset of $\mathbb{Z}^2$: it is a grid of points (keys). Here we look to embed our 26 letters in a $3 \times 9 = 27$-key keyboard.
\medskip

\subsection{First problem}

The original problem of minimizing the long-run distance covered by the hand can be defined as the following IP problem. Here, $\pi_i$ is the long-run probability of typing $i$, $f_{ij}$ is the frequency of the transition $i \rightarrow j$, $d_{kl}$ is the distance between key $k$ to key $l$ (on the keyboard) and $x_{ij}$ is 1 if letter $i$ is matched to key $j$ and 0 otherwise:

$$\text{minimize}_x\, \sum_i \sum_j \sum_k \sum_l \pi_i f_{ij}d_{kl}x_{ik}x_{jl}$$
$$s.t. 
\begin{cases} \sum_i x_{ij} \leq 1 \text{ (every key is linked to at most 1 letter)} $$ \\
$$\sum_j x_{ij} = 1 \text{ (every letter is linked to one key)} $$  \\
$$x_{ij} \in \{0, 1\} \end{cases}$$

We tried to implement this optimization problem with the \textit{gekko} and \textit{scikit} libraries, however it is far too complex to be tractable. Splitting the alphabet into the two clusters that we found, and thereby drastically reducing the number of variables, does not seem to improve the performance whatsoever. 

In fact, numerous studies have tried to solve this type of integer problem and most of them use much more complex algorithms such as deep genetic algorithms, which we feel is not really relevant to our original objective of using geometric tools to uncover the optimal keyboard. We will then use simpler approximations.

\subsection{Making use of the 2D embedding}

Instead of starting new with integer programming, we can also leverage the 2D-embedding of the alphabet that we found in the above section, and look for the optimal keyboard layout from then only. 

In that regard, the layout that preserves the distances of the 2D-embedding as well as possible should result in a keyboard optimal enough.
Therefore, we want to find the optimal isometric (as much as possible) embedding that minimizes the Gromov-Hausdorff distance between the keyboard $Y$ and the 2D-embedding of our points, that we note $X$. Both $X$ and $Y$ are endowed with the Euclidean metric. With $m$ the corresponding (letter $\rightarrow$ key) injective matching from $X$ to $Y$, this reads:

$$\text{minimize}_m \, d_{GH}(X,m(X)) = \frac{1}{2} \inf_{\varphi \text{ isometry}} \sum_{x \in \varphi(X)} ||x - m(x)||_2^2$$

We approach this problem the other way around. To do so, we try and compute an isometry of $X$ that best estimates the shape of the keyboard embedding, and then we will assess the optimal matching. For that first part, we scower through simple linear isometries: combinations of translations and rotations.

In the first place, as the keyboard has not been scaled yet, we normalize both $X$ and $Y$. Then we center them with simple translations, before computing the rotation that aligns both sets of points at best. To do so we compute the angle that aligns the covariance matrices by making them diagonal. Since $Var(OX) = O^TVar(X)O$ and we want to have $Var(OX)$ diagonal, aligning both covariance matrices is equivalent to finding the eigenvectors of $Var(X)$ (which is symmetric positive so diagonalizable in an orthogonal basis) and applying the rotation to $X$.

The following is an example of rotating the first cluster for 2-handed typing to best approach the shape of the keyboard.

\begin{figure}[H]
  \centering
  \includegraphics[scale=.6]{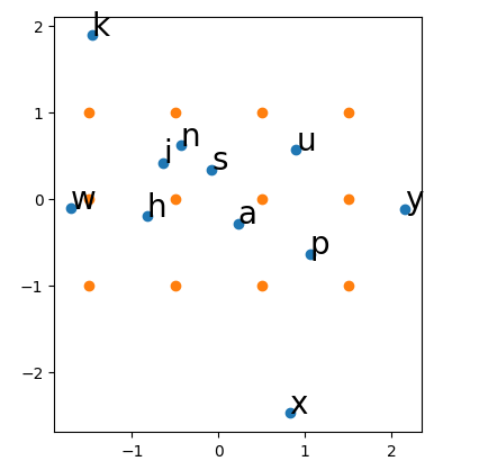}
  \caption{After scaling, centering, before rotation.}
\end{figure}

\begin{figure}[H]
  \centering
  \includegraphics[scale=.6]{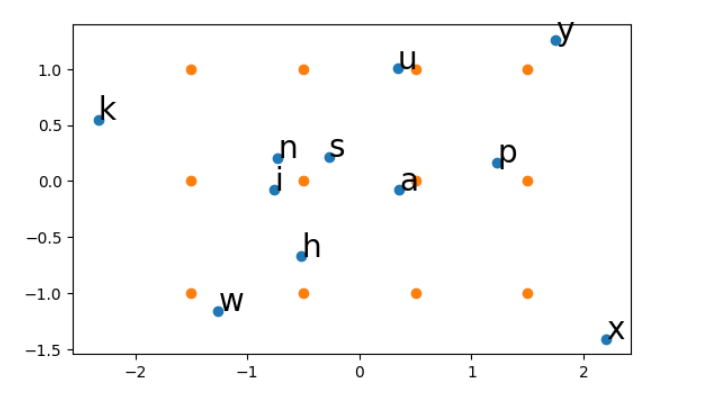}
  \caption{After rotation.}
\end{figure}

\subsubsection{Computing the matching: sorting keys}

Then, since $Y$ is a grid of $3$ rows of $9$ columns, we can split the points by $y$ coordinates by bins of 9, and each bin is then sorted by $x$ coordinate, which gives in the end the keyboard layout. 

\begin{center}
\includegraphics[scale=1]{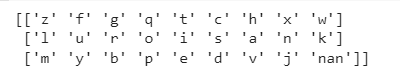}
\end{center}

Finally, we can apply the same method for both clusters (that we have found in the previous part). Since the clusters are of size $11$ and $15$, we put in $3 \times5$ sub-keyboard on the left, that we merge with a $3 \times4$ sub-keyboard on the right and we obtain the following result : 

\begin{center}
\includegraphics[scale=1]{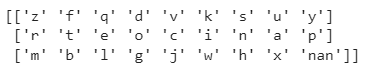}
\end{center}

However, even though this method should give sensible results, it does not take into account any notion of minimizing distance: sorting the points does not result in a good enough keyboard.

\subsubsection{Computing the matching: Integer Programming}

One other way to compute the matching is to transform the above problem into an integer programming one, now that we have approached $\varphi$. With $x_{ij}$ being the matching variable between letter $i$ $l_i$ and key $j$ $k_j$, this reads:

$$\text{minimize}_x \, \sum_{ij} x_{ij} ||l_i - k_j||_2^2$$
$$s.t.
\begin{cases}
    \sum_i x_{ij} \leq 1 \text{ (every key is linked to at most 1 letter)} $$ \\
$$\sum_j x_{ij} = 1 \text{ (every letter is linked to one key)} $$  \\
$$x_{ij} \in \{0, 1\} \\
\end{cases}$$

This integer linear problem can be easily solved, as the integer condition can be relaxed. The solution of the ILP is the same as the one of the LP, because the constraint matrix $A$ is unimodular, as it respects a sufficient condition : $A \in \{0,1\}^{52\times 26^2},\, \sum_i A_{ij} = 2,\, A = (B,C)^T$ with $B$ being the conditions on $i$, $C$ the conditions on $j$.

Hence the Simplex algorithm can solve this problem in polynomial time. We get the following optimal keyboard:

\begin{figure}[h!]
    \centering
    \includegraphics[scale=.9]{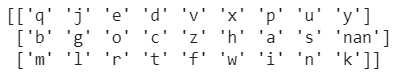}
    \caption{The optimal H2 keyboard}
\end{figure}

\newpage
{\color{gray}\hrule}
\begin{center}
\section{Validation of the model}
\end{center}

{\color{gray}\hrule}

\vspace{10pt}

To verify and solve the theoretical problem, we implemented our approach with a mix of optimization and geometric/graphical tools. The first step is to code the optimization problem for the H1 keyboard, then the max cut refines the problem for the H2 keyboard. In this logic, we set up a geometric transformation to transform the embedding we have obtained into a keyboard. Finally, we carry out geometric analysis to assess the underlying patterns of our keyboards.

\medskip

\subsection{Testing framework}

We want to prove here that the keyboard we have obtained reduces the time we spend typing. If necessary, we re-scale it in a first place so that the distance between adjacent keys is the same as a similar QWERTY layout. Then, we construct a framework to compare the two typing styles. For the following, we assume that the speed of hand is constant. We assume that the typing time is unimportant in our case, mainly because the keyboards we compare share the same one (it depends on the manufacturing of the keyboard, not on the layout). As we only want to compare relative typing speeds, this does not seem relevant in the context of our study.

The text we use to carry out the testing is sampled from the book \underline{Alice in Wonderland} by Lewis Carroll, found at \href{https://www.gutenberg.org/ebooks/11}{Project Gutenberg}. For testing purposes, we have kept the main content only, we have removed all possible punctuation marks (including spaces and newlines), and have transformed all letters to lowercase. The resulting test set is, in a sense, a single string of $104174$ lowercase letters.

Our testing is done in terms of arbitrary units (a.u.), given there isn't a need to scale the keyboards.

\subsubsection{Results for the H1 keyboard}
From the QWERTY keyboard and the 1-handed keyboard H1 we extract the distance matrices $D_{QWERTY-1}$ and $D_1$. We do not take the space bar into account. Given a text, we simply compute the distance covered by the hand to write the text.
The results are the following:

\begin{figure}[h!]
    \centering
    \includegraphics[scale=.4]{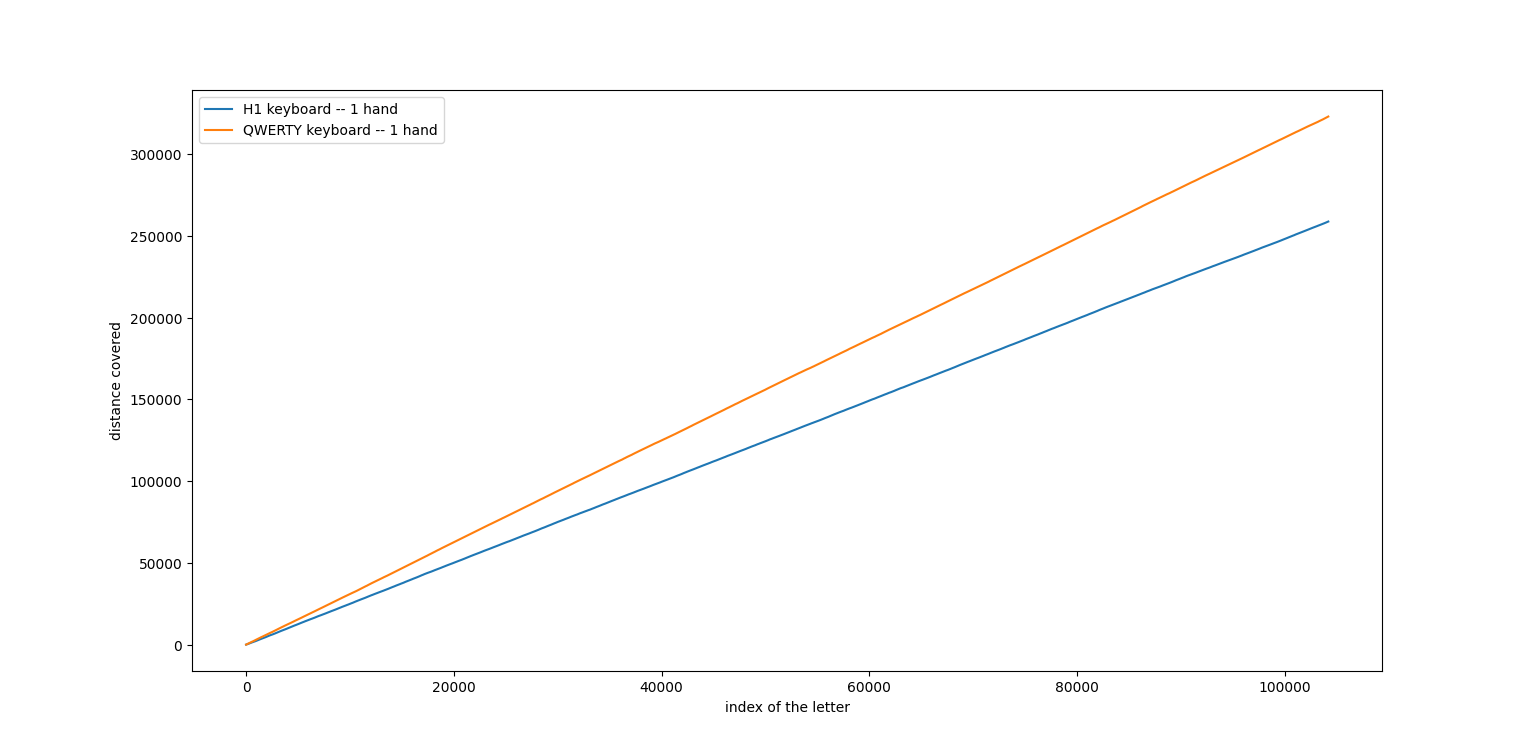}
    \caption{Comparison of the performances of H1 and QWERTY-1 keyboards.}
\end{figure}

We observe that the H1 keyboard covers $2.48$ a.u. per transition while the QWERTY-1 covers $3.10$ a.u.. This is a $20\%$ improvement in the performance of our keyboard.

\subsubsection{Results for the H2 keyboard}
For the 2-handed keyboard H2, we cluster the QWERTY keyboard in 2 groups of letters: all letters above or on the left of \textit{b}, and the rest. Measuring the time spent typing is a little trickier here, because the distance covered by one hand does not necessarily entail more total writing time. \\
Assume we are in the previous example, where the keyboard has the letters \textit{a} and \textit{w} in the left keyboard, and \textit{n, d, r, e} in the right keyboard. Then, when writing \textit{andrew}, the left hand will first type \textit{a}, and then go straight to \textit{w}, where it may 'wait' for the right hand to go through \textit{n, d, r, e}, before typing \textit{w}. If the left hand does wait (i.e., if $d_{aw} \leq d_{nd}+d_{dr}+d_{re}$), then the transition time from \textit{e} to \textit{w} is negligible, and we consider here that it is actually instant (we will only account for the typing of \textit{w}). Else (if $d_{aw} > d_{nd}+d_{dr}+d_{re}$), the extra time spent is reflected in the quantity $d_{aw} - d_{nd} + d_{dr} + d_{re}$.

\begin{figure}[h!]
    \centering
    \includegraphics[scale=.4]{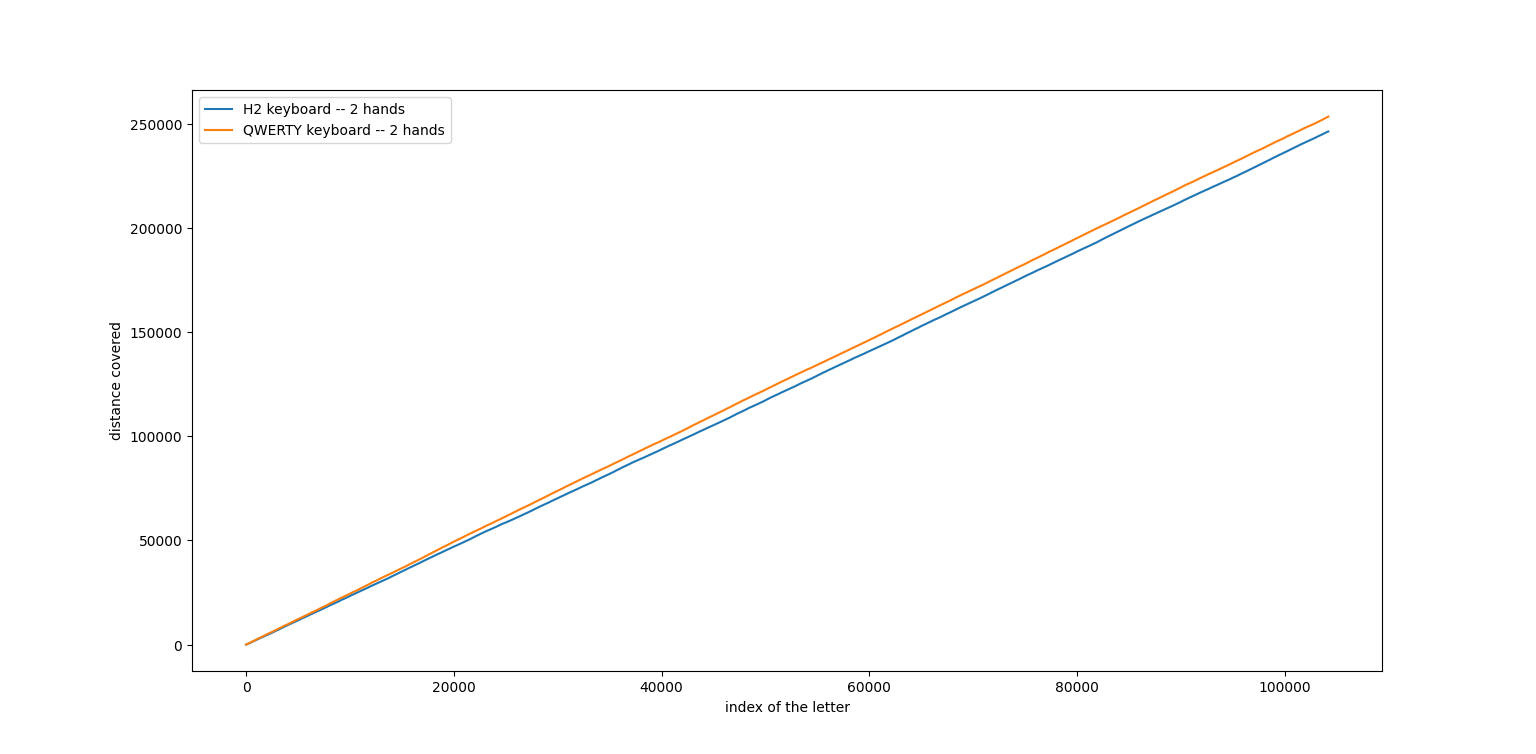}
    \caption{Comparison of the performances of H2 and QWERTY-2 keyboards}
\end{figure}

We observe that the H2 keyboard covers $2.36$ a.u. per transition while the QWERTY-2 covers $2.43$ a.u.. This is a $2.8\%$ improvement in the performance of our keyboard. The results are not as conclusive as expected, but this is still a non-negligible improvement. It turns out the QWERTY keyboard does reduces the long-run distance covered by the hands pretty well already.

\subsection{Covariance analysis}
To validate our experiment and the efficiency of our keyboard, we plot the covariance ellipses of both our keyboards and the QWERTY: they reflect on the dispersion of the hand throughout the keyboards. This allows to assess the differences in usages for the keyboards.\\

\textbf{Definition :} Let $\mu$ be the $\pi$-weighted centroid of our space $X\in \mathbb R^{26\times 2}$, $\Sigma$ the $\pi$-weighted covariance matrix, $P$ the rotation matrix from the canonical basis to $\Sigma$ spectral basis.

We define by $(x-\mu)^TP\Sigma^{-1}P^T(x-\mu) = 1$ the equation of the covariance ellipse.\\

Plotting covariance ellipse for keyboards gives geometric interpretable results. We can see the dispersion of the distance we travel in average, from one touch to the other in a cluster. The experiment, on\textbf{ Figure 11}, totally confirms our model:

\begin{figure}[h!]
  \centering
  \begin{subfigure}[b]{0.45\textwidth}
    \includegraphics[width=\linewidth]{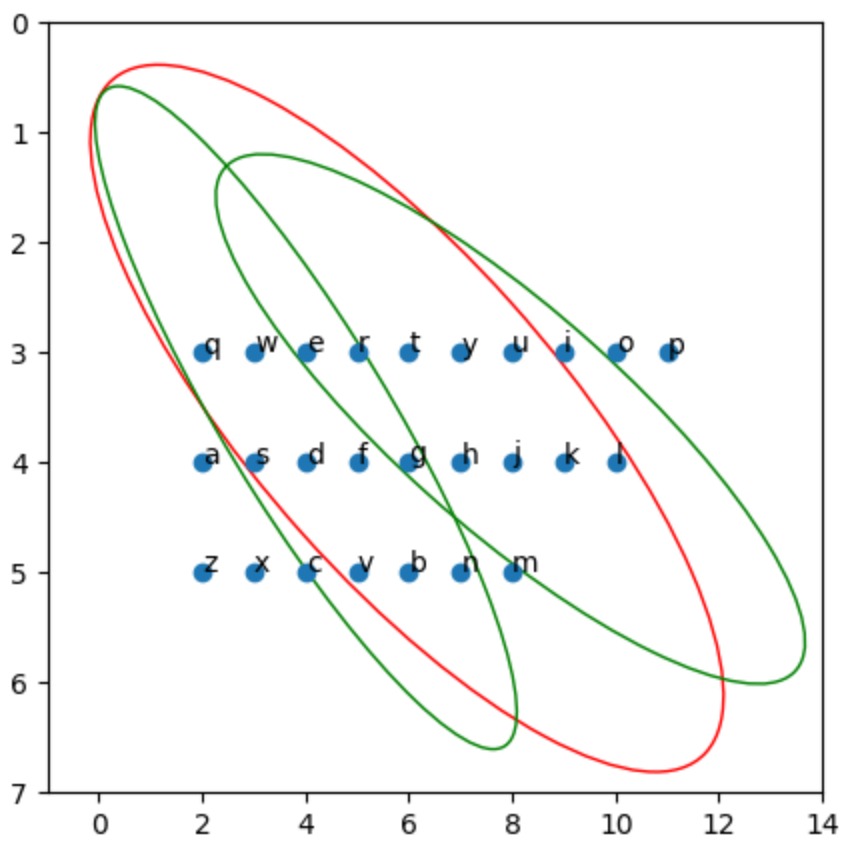}
    \caption{QWERTY keyboard}
    \label{fig:image1}
  \end{subfigure}
  \hfill
  \begin{subfigure}[b]{0.45\textwidth}
    \includegraphics[width=\linewidth]{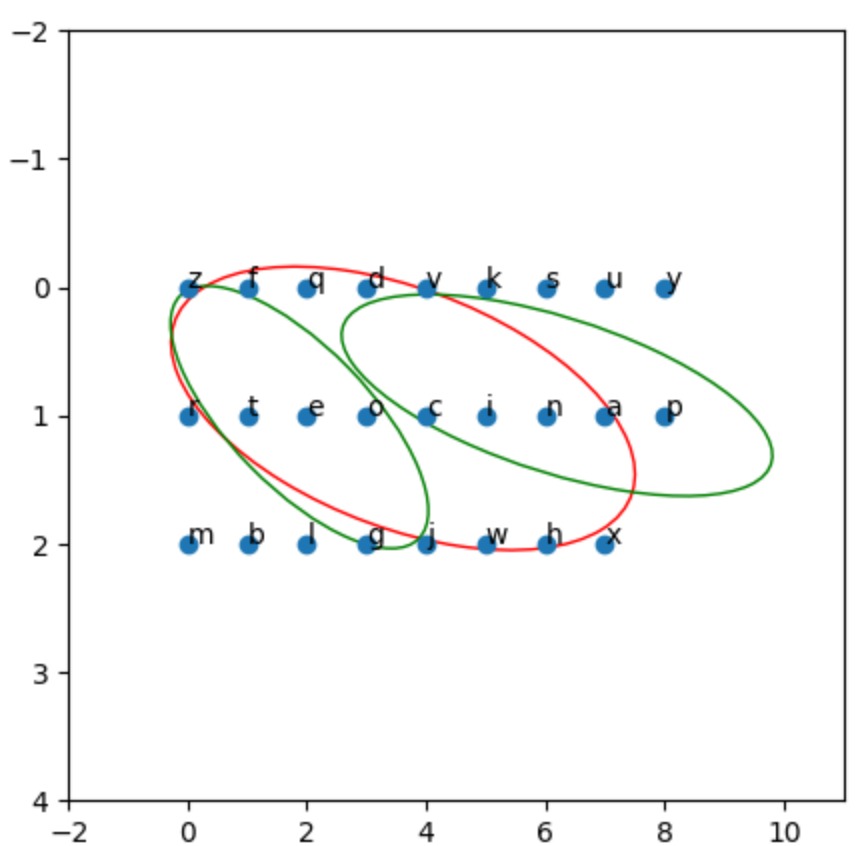}
    \caption{H2 keyboard}
    \label{fig:image2}
  \end{subfigure}
  \caption{Covariance ellipse for the H2-keyboard, in green: the ellipse for each, in red: the entire ellipse}
  \label{fig:overall}
\end{figure}

We notice, as expected, that the distance covered is much more concentrated in our H2-keyboard than in the QWERTY, which means the latter should be less efficient, as the correlated keys, i.e. those we type one just after the other, are far.

\subsection{Limitations}

While the novel keyboards we propose offer some advantages over the QWERTY keyboard, there are still many improvements to be made in our study:
\begin{itemize}
    \item keeping track of the distribution of the space bar throughout the English language (i.e., how words begin and end), and all punctuation at large, might change totally the layout of the keyboard, and seems to be essential in the design of an efficient kayboard layout;
    \item the hypothesis that typing with 2 hands is equivalent to partitioning the keyboard into 2 clusters is highly debatable;
    \item the hyperparameters for the optimization problem might need refining.
\end{itemize}

\newpage

{\color{gray}\hrule}
\begin{center}
\section{Openings}
\end{center}
{\color{gray}\hrule}

\vspace{10pt}

\subsection{International optimized keyboard}
\subsubsection{Definition of the problem}
To improve the relevance of the keyboard, it should be optimal for people using several languages. To reach this goal, we can think of a keyboard designed from words of several languages. 

Our approach is to use datasets of frequently used words in different languages and to compute a frequency matrix of sequences of two letters for each language. The rows of a frequency matrix can be understood as conditional probability distributions, as they are Markov Transitions Probabilities. Hence, to compute an average of the frequency, the Wasserstein distance is totally relevant, as it allows to define a barycenter between all these language distributions, which acts as an "average Markov Chain". This barycenter captures geometric properties of the language distributions, and especially the curvature, which is for the Keyboard Problem approximately the letter stationary distribution.

By writing $P_i$ the transition probability matrix in language $i$, and after normalizing the distributions, the problem becomes :
$$P_{barycenter} = \text{argmin}_P \sum_i d_W(P_i,P)^2 *C$$
with $$d_W(p,q) = \inf_{\pi \text{ coupling of p,q}}\int ||x-y||d\pi(x,y)$$
an optimal transport distance and $C$ a normalization constant for each line.

It only remains to do the optimization process described before to get the "average" optimal keyboard, in a geometric sense.

\subsubsection{Analysis of keyboards}
As part of the making of an international keyboard, it is interesting to study the differences between keyboards from different origins. For that purpose we can assess how much keyboards distort from one another.

It is well known that the Gromov-Hausdorf distance measures the distortion between two spaces. In order to carry out our analysis, we decided to create a close metric. We remind that under usual manifold hypothesis, $d_{GH} = \max_{(x,y),(x',y')\in C} |d_X(x,x')-d_Y(y,y')|$. Here this is a max given the finitude of the sets of points.

If we decide to identify each key with its letter, the most intuitive coupling between the two keyboards is $f : x_{keyboard} \longrightarrow y_{qwerty}$ which associates to the position of a letter in our keyboard the position of this one in the QWERTY.

A one-hand model gives a home-made distance : $d_{GH} = \max_{(x,y),(x',y') \in C} |d(x,x')-d(y,y')|$, with $d$ the Euclidian metric, as our hands travel this distance, which represents the biggest relative distortion between the two keyboard. To refine this distance, we can ponder the points with the stationary distribution, to add importance to the most important letters.

\newpage

{\color{gray}\hrule}
\begin{center}
\section{Conclusions}
\end{center}
{\color{gray}\hrule}

\vspace{0.5cm}

In conclusion, the framework that we propose to design more efficient keyboard layouts does give some interesting results: the H1 keyboard is $20\%$ faster than QWERTY-1, and the H2 keyboard is $3\%$ faster than QWERTY-2. We understand that optimizing 2-handed typing is much harder than 1-handed, or that the QWERTY keyboard does pretty well at this job already. However, the diversity in the approaches that we have taken here provides us with a solid basis to continue building upon: fine-tuning the optimization problems, and perhaps leveraging better quantities such as the curvature of the keyboard, should allow us to develop an even more performant keyboard layout in the end.\\

A Python notebook showcasing the main outline of this study is accessible \href{https://colab.research.google.com/drive/1r_Zr-VMguBd8icFakin48SCRVZYW-dcX?usp=sharing}{here}.
\bibliographystyle{plain}
\bibliography{./references}

\begin{thebibliography}{1}

\bibitem{ant2003}
Jan~Eggers et~al.
\newblock Optimization of the keyboard arrangement problem using an ant colony
  algorithm.
\newblock {\em European Journal of Operational Research 148}, 2003.

\bibitem{Noyes1982}
Jan Noyes.
\newblock The qwerty keyboard: a review.
\newblock {\em International Journal of Man-Machine Studies}, 1982.

\bibitem{genetic2020}
Amir Hosein~Habibi Onsorodi and Orhan Korhan.
\newblock Application of a genetic algorithm to the keyboard layout problem.
\newblock {\em PLoS ONE 15}, 2020.

\bibitem{swarm2011}
Peng-Yeng Yin and En-Ping Su.
\newblock Cyber swarm optimization for general keyboard arrangement problem.
\newblock {\em International Journal of Industrial Ergonomics}, 2011.

\end{thebibliography}

\end{document}